\documentclass[conference]{IEEEtran}
\IEEEoverridecommandlockouts
\usepackage[english]{babel}
\usepackage{cite}
\usepackage{float}
\usepackage{cuted}
\usepackage{amsthm}
\usepackage{amsmath,amssymb,amsfonts}
\usepackage{hyperref}
\usepackage{array}
\usepackage{algpseudocode}
\usepackage{algorithm}
\usepackage{graphicx}
\usepackage{textcomp}
\usepackage{xcolor}
\usepackage{indentfirst}
\usepackage{booktabs}
\def\BibTeX{{\rm B\kern-.05em{\sc i\kern-.025em b}\kern-.08em
    T\kern-.1667em\lower.7ex\hbox{E}\kern-.125emX}}
    
\newtheorem{assumption}{Assumption}
\newtheorem{definition}{Definition}
\newtheorem{theorem}{Theorem}
\newtheorem{corollary}{Corollary}[theorem]

\newtheorem{lemma}[theorem]{Lemma}

\newcommand{\SIHT}{\textsc{Sto}IHT}
\newcommand{\STO}{\textsc{GraphSto}-IHT}
\newcommand{\SVRG}{\textsc{Graph}SVRG-IHT}
\newcommand{\SCSG}{\textsc{Graph}SCSG-IHT}

\newcommand{\bb}[1]{\mathbb{#1}}
\newcommand{\K}{\mathcal{K}}
\newcommand{\tpm}{\textpm}

\begin{document}

\title{Stochastic Variance-Reduced Iterative Hard Thresholding in Graph Sparsity Optimization\\
}

\author{\IEEEauthorblockN{Derek Fox}
\IEEEauthorblockA{\textit{Computer Science} \\
\textit{UNC Greensboro, USA}\\
defox@uncg.edu}
\and
\IEEEauthorblockN{Samuel Hernandez}
\IEEEauthorblockA{\textit{Mathematics} \\
\textit{Texas A\&M University, USA}\\
samuelhq11@tamu.edu}
\and
\IEEEauthorblockN{Qianqian Tong}
\IEEEauthorblockA{\textit{Computer Science} \\
\textit{UNC Greensboron, USA}\\
q\_tong@uncg.edu}
}

\maketitle

\begin{abstract}
  Stochastic optimization algorithms are widely used for large-scale data analysis due to their low per-iteration costs,  but they often suffer from slow asymptotic convergence caused by inherent variance. Variance-reduced techniques have been therefore used to address this issue in structured sparse models utilizing sparsity-inducing norms or $\ell_0$-norms. However, these techniques are not directly applicable to complex (non-convex) graph sparsity models, which are essential in applications like disease outbreak monitoring and social network analysis. In this paper, we introduce two stochastic variance-reduced gradient-based methods to solve graph sparsity optimization: \textsc{Graph}SVRG-IHT and \textsc{Graph}SCSG-IHT. We provide a general framework for theoretical analysis, demonstrating that our methods enjoy a linear convergence speed. Extensive experiments validate the efficiency and effectiveness of our proposed algorithms.
\end{abstract}

\section{Introduction}

\noindent Graph structures enable the imposition of intricate sparsity constraints on the model, allowing them to better reflect relationships present in the data. For instance, graph-structured sparsity models are well-suited for predicting the spread of diseases or identifying social groups in networks. The search of connected subgraphs or clusters has a significant impact on identifying disease-related genes \cite{arias2011detection, qian2014connected, aksoylar2017connected, rozenshtein2014event}. Graph sparsification, which aims to reduce the complexity of large-scale graphs while preserving their essential structural properties, has garnered increasing attention as a crucial technique in modern data analysis and machine learning \cite{baraniuk2010model,hegde2014fast, hegde2015approximation, hegde2015nearly, zhou2019stochastic}. 

Graph sparsification can be formulated as the following optimization problem:
\begin{equation} \label{eq:1}
\min_{x \in \mathbb{R}^p} F(x), \quad F(x) := \frac{1}{n} \sum_{i=1}^{n} f_i(x),
\end{equation}
which is known as the empirical risk minimization problem. Each $f_i(x)$ $(i\in [n] )$ is convex and differentiable, and the graph 
structured sparsity is reflected by the constraint set $\mathbb{R}^p$ on $x$.  The input vector $x$ denotes the parameter of the model, and the output $f_i(x)$ is defined as the loss associated with sample $i$. By minimizing the loss $F(x)$, we guide the model towards the optimal solution. Typically, sparsity can be encoded by adding sparsity-inducing norms or penalties such as $\ell_0$ norm, $\ell_1$ norm and mixed norms \cite{blumensath2009iterative, foucart2011hard, yuan2014gradient, jain2014iterative, tibshirani1996regression, van2008high, chen2001atomic, turlach2005simultaneous, yuan2006model}.  These models often involve convex penalties and can be solved using convex optimization algorithms \cite{bach2012optimization, bach2012structured, bahmani2013greedy}. However, dealing with more complex sparse settings, such as graph-structured models, is more challenging.

In stochastic optimization, iterative hard thresholding (IHT) methods include gradient descent IHT (GD-IHT) \cite{jain2014iterative}, stochastic gradient descent IHT (SGD-IHT) \cite{nguyen2017linear}, hybrid stochastic gradient IHT (HSG-IHT) \cite{zhou2018efficient}, and variance-reduced methods such as stochastic variance reduced gradient IHT (SVRG-IHT) \cite{li2016stochastic}, stochastically controlled stochastic gradient IHT (SCSG-IHT) \cite{liang2020effective}.
These methods update the parameter iterate $x$ via gradient descent or its variants, and then apply a hard thresholding (HT) operator to enforce sparsity of $x$, preserving the top $s$ elements in $x$ while setting other elements to zero.
In the context of graph-structured sparse optimization, the stochastic gradient decent based IHT method, named \STO, achieves HT through the use of Head and Tail Projections, first described by \cite{zhou2019stochastic}. Head and Tail Projections map arbitrary vectors from the data onto the graph while simultaneously enforcing model sparsity \cite{hegde2015approximation, hegde2015nearly, hegde2016fast}. Specifically, Head Projection identifies and preserves the largest entries in $x$, while Tail Projection identifies the smallest entries and sets them to zero. By ignoring the small magnitude entries in the vector, these projections help prevent overfitting and ensure sparse solutions.
Meanwhile, stochastic sampling of the data is used to speed up gradient calculations. A single data point or small batch of the data is selected and a gradient is calculated only with respect to that batch. This greatly decreases the computational costs associated with the gradient calculations. However, if the selected batch  does not represent the whole dataset, the gradient may not accurately point towards a local minimum of the function, introducing variance into the gradient descent process.

To reduce the randomness inherent in SGD, variance reduction techniques may be used such as SVRG \cite{johnson2013accelerating}, SAGA \cite{defazio2014saga}, or SCSG \cite{lei2017less}. During each iteration, the history of the stochastic process is considered to regulate the newly calculated gradient and minimize large changes in direction. This improvement on SGD is our main interest; both proposed algorithms utilize this technique to leverage fast gradient calculations while still enjoying quick convergence. In this paper we leverage the recent success of stochastic variance-reduced algorithms for non-convex problems and propose a series of efficient stochastic optimization algorithms for graph-structured sparsity constraint problems. Specifically, we introduce two new stochastic variance-reduced gradient based methods, \SVRG\ and \SCSG, designed to solve graph sparsity optimization problems. By incorporating stochastic variance-reduced techniques and graph approximated projections (head and tail), our algorithms are specifically tailored for non-convex graph-structured sparsity constraints, leading to faster convergence and improved performance. We provide a comprehensive theoretical framework and conduct extensive experiments to validate the efficiency and effectiveness of our proposed methods.

\textbf{Our main contributions} are summarized as follows.
\begin{itemize}
    \item This work is the first to explore the application of stochastic variance-reduced methods to graph-structured sparsity problems. By using batch gradients to approximate computationally expensive full gradients, we enhance the efficiency of variance reduction. \SCSG\ is built on \SVRG\ by parameterizing the variance reduction technique for greater control. It employs two different batch sizes for gradient calculations and randomly selects the update rate of the current position from a geometric distribution.
 
    \item  We theoretically prove \SCSG\ enjoys linear convergence rate with a constant learning rate.  This analysis provides a robust theoretical framework for analyzing graph sparsification optimization.
    
    \item We conduct extensive experiments to validate our proposed algorithms. In addition to simulation tests, we also evaluate our methods on a real-world breast cancer dataset. Our experiments empirically demonstrate the efficiency and effectiveness of our methods compared to \STO\ and other deterministic methods.
\end{itemize}

\section{Preliminaries}

\noindent \textbf{Notations.} We use lowercase letters, e.g. $x$, to denote a vector and use $\|\cdot\|$ to denote the $l_2$-norm of a vector. The operator $E$ represents taking expectation over all random variables, $[n]$ denotes the integer set $\{1, ..., n\}$. The notation
$supp( x )$ means the support of $x$ or the index set of non-zero elements in $x$.  Other important parameters are listed in Appendix \ref{table:parameters}.

\begin{definition} (Subspace model) \cite{hegde2016fast}
\label{def:subspace}
 Given the space \(\mathbb{R}^p\), a subspace model \(\mathcal{M}\) is defined as a family of linear subspaces of \(\mathbb{R}^p\):
\[
\mathcal{M} = \{S_1, S_2, \ldots, S_k, \ldots \}
\]
where each \(S_k\) is a subspace of \(\mathbb{R}^p\). The set of corresponding vectors in these subspaces is denoted as
\[
\mathcal{M}(\mathcal{M}) = \{x : x \in V \text{ for some } V \in \mathcal{M}\}.
\]

\end{definition}

\begin{definition} (Weighted graph model) \cite{hegde2015nearly}
\label{def:weighted}
Given an underlying graph \(G = (V, E)\) defined on the
coefficients of the unknown vector \(x\), where \(V = [p]\) and
\(E \subseteq V \times V\), then the weighted graph model \((G, s, g, C)\)-
WGM can be defined as the following set of supports
\[
\mathcal{M} = \{S : |S| \leq s, \text{ there is an } F \subseteq V \text{ with } 
\]
\[V_F = S, \gamma(F) = g, \text{ and } w(F) \leq C\},
\]
where \(C\) is the budget on weight of edges \(w\), \(g\) is the number
of connected components of \(F\), and \(s\) is the sparsity.
\end{definition}

\begin{definition}(Projection Operator) \cite{hegde2016fast}
\label{def:projop}
    We define a projection operator onto $\mathcal{M}(\mathbb{M})$, i.e, $P(\cdot, \mathcal{M}(\mathbb{M})):\mathbb{R}^p \to \mathbb{R}^p$ defined as
    \begin{equation}
        P(x, \mathcal{M}(\mathbb{M})) = \arg\min_{y \in \mathcal{M}(\mathcal{M})} \|x - y\|^2. \notag
    \end{equation}
\end{definition}

With this projection definition, we borrow two important projections: Head Projection and Tail Projection from previous literature to help us with theoretical analysis.

\begin{assumption} (Head Projection)\cite{hegde2016fast}
\label{ass:head}
     Let $M$ and $M_H$ be the predefined subspace models. Given any vector $x$, there exists a $(c_H, M, M_H)$-Head-Projection which is to find a subspace $H \in M_H$ such that
\[
\|P(x, H)\|_2 \geq c_H \cdot \max_{S \in M} \|P(x, S)\|_2, \tag{1}
\]
where $0 < c_H \leq 1$. We denote $P(x, H)$ as $P(x, M, M_H)$.
\end{assumption}

\begin{assumption} (Tail Projection)\cite{hegde2016fast}
\label{ass:tail}
     Let $M$ and $M_T$ be the predefined subspace models. Given any vector $x$, there exists a $(c_T, M, M_T)$-Tail-Projection which is to find a subspace $T \in M_T$ such that
\[
\|P(x, T) - x\|_2 \leq c_T \cdot \min_{S \in M} \|x - P(x, S)\|_2, \tag{2}
\]
where $c_T \geq 1$. We denote $P(x, T)$ as $P(x, M, M_T)$.

\end{assumption}

We can see that head projection keeps large magnitudes whereas tail projection sets small magnitudes to zero.

\begin{definition} (($\alpha, \beta, M(M)$)-RSC/RSS Properties)\cite{hegde2016fast}
\label{def:RSC/RSS}
    We say a differentiable function $f(\cdot)$ satisfies the $(\alpha, \beta, M(M))$-Restricted Strong Convexity (RSC)/Smoothness (RSS) property if there exist positive constants $\alpha$ and $\beta$ such that
\[
\frac{\alpha}{2} \|x - y\|_2^2 \leq B_f(x, y) \leq \frac{\beta}{2} \|x - y\|_2^2, \tag{3}
\]
for all $x, y \in M(M)$, where $B_f(x, y)$ is the Bregman divergence of $f$, i.e.,
\[
B_f(x, y) = f(x) - f(y) - \langle \nabla f(y), x - y \rangle.
\]
$\alpha$ and $\beta$ are the strong convexity parameter and strong smoothness parameter, respectively.
\end{definition}

The RSC/RSS definition is widely used in sparsity optimization. Together with the following assumption, we characterize the properties of the objective function.

\begin{assumption} \cite{hegde2016fast}
\label{ass:three}
    Given the objective function \(F(x)\) in (1), we assume that \(F(x)\) satisfies \(\alpha\)-RSC in subspace model \(\mathcal{M}(\mathcal{M} \oplus \mathcal{M}_H \oplus \mathcal{M}_T)\). Each function \(f_i(x)\) satisfies \(\beta\)-RSS in \(\mathcal{M}(\mathcal{M} \oplus \mathcal{M}_H \oplus \mathcal{M}_T)\), where \(\oplus\) of two models \(\mathcal{M}_1\) and \(\mathcal{M}_2\) is defined as \(\mathcal{M}_1 \oplus \mathcal{M}_2 := \{S_1 \cup S_2 : S_1 \in \mathcal{M}_1, S_2 \in \mathcal{M}_2\}\).

\end{assumption}

\section{Methods}

\noindent In this section, we introduce two proposed algorithms: \SVRG\ and \SCSG. Both algorithms employ the variance-reduced techniques derived from SVRG \cite{johnson2013accelerating} and SCSG \cite{lei2017less} respectively, while also utilizing the graph projection operators found in \STO \cite{zhou2019stochastic}. This results in methods that are applicable to graph-structured sparsity problems and effectively reduce the variance inherent to stochastic gradient descent. Therefore, our algorithms converge faster and more accurately than their predecessors.

\subsection{\SVRG}

\noindent  
Our proposed \SVRG\ algorithm (Algorithm \ref{alg:graph_svrg_iht}) utilizes variance reduction by periodically computing the full gradient,  significantly reducing the inherent variance in stochastic gradient methods. By incorporating graph projection operators, our \SVRG\ adapts to non-convex graph sparsity constraints, enhancing its applicability and efficiency. The key steps are outlined below:

\begin{enumerate}
    \item Calculate the full gradient, $\tilde{v}^j$, with the position at the start of each outer loop, $\tilde{x}^j$ (Line 4). 
    \item In the inner loop, compute two gradients from a single sampled data point: one at the copied position $x^j_k$ and the other at $\tilde{x}^j$. Then calculate the stochastic variance reduced gradient, $v_k^j$ (Line 7-8).
    \item  Pass $v_k^j$ through the Head Projection operator (Line 9); and use the resulting gradient to update the next iterate $x^j_k$, through the Tail Projection operator (Line 10).
    \item After a fixed number ($\mathcal{K}$) of inner loop iterations, update the outer loop position $\tilde{x}^j$ and re-calculate the new full gradient.
\end{enumerate}

\begin{algorithm}[h]
\caption{\SVRG}\label{alg:graph_svrg_iht}
\begin{algorithmic}[1]
    \State \textbf{Input}: $\eta$, $f_i$, $\mathbb{M}$, $\mathbb{M}_\mathcal{H}$, $\mathbb{M}_\mathcal{T}$, $\mathcal{J}$, $\mathcal{K}$
    \State \textbf{Initialize}: $\tilde{x}^1$ such that $supp(\tilde{x}^1)\in \mathbb{M}$
    \For {$j = 1$ to $\mathcal{J}$}
        \State $\tilde{v}^j = \frac{1}{n} \sum_{i=1}^n{\nabla f_i(\tilde{x}^j)}$
        \State $x^j_0 = \tilde{x}^j$
        \For{$k = 1$ to $\mathcal{K}$}
            \State Randomly pick $i_k \in [n]$
            \State $v^j_k = \nabla f_{i_k}(x^j_{k-1}) - \nabla f_{i_k}(\tilde{x}^j) + \tilde{v}^j$
            \State $\tau_k = P(v^j_k, \mathbb{M} \oplus \mathbb{M}_\mathcal{T}, \mathbb{M}_\mathcal{H})$
            \State $x^j_k = P(x^j_{k-1} - \eta \tau_k, \mathbb{M}, \mathbb{M}_\mathcal{T})$
        \EndFor
        \State $\tilde{x}^{j+1} = x^j_\mathcal{K}$
    \EndFor
\end{algorithmic}
\end{algorithm}

Our proposed \SVRG\ algorithm differs from \STO\ in that \SVRG\  uses nested loops, allowing it to account for the history of the stochastic process, whereas \STO\ only considers a stochastic gradient. Both methods implement Head and Tail Projections for hard thresholding, making them applicable to graph-structured sparsity optimization problems. The inclusion of stochastic variance in \SVRG\ makes the theoretical analysis more complex and challenging.

\subsection{\SCSG}
\noindent To better understand the calculation of variance-reduced gradients and stochastically control the outer batch size, we propose the \SCSG\ algorithm (Algorithm \ref{alg:graph_scsg_iht}). While similar to  Algorithm \ref{alg:graph_svrg_iht} in its use of variance-reduced gradients, \SCSG\ has the following key characteristics:

\begin{enumerate}
    \item In the outer loop, the gradient is calculated using a batch of data of size $B$, whereas Algorithm \ref{alg:graph_svrg_iht} calculates a full gradient at this step (Line 4-5).
    \item In the inner loop, when calculating the stochastic variance reduced gradient, a mini-batch is used instead of a single data point (Line 10-11).
    \item The number of inner loops, $\mathcal{K}^j$, is not fixed. Instead, $\mathcal{K}^j$ is chosen from a geometric distribution (Line 7) or can be set as $\frac{B}{b}$ (Line 8).
\end{enumerate}

\begin{algorithm}[h]
\caption{\SCSG}\label{alg:graph_scsg_iht}
\begin{algorithmic}[1]
\State \textbf{Input}: $\eta$, $f_i$, $\mathbb{M}$, $\mathbb{M}_\mathcal{H}$, $\mathbb{M}_\mathcal{T}$, B, b, $\mathcal{J}$
\State \textbf{Initialize}: $\tilde{x}^1$ such that $supp(\tilde{x}^1)\in \mathbb{M}$
\For{$j = 1$ to $\mathcal{J}$}
    \State Uniformly sample a batch $I^j \subset [n]$, s.t. $|I^j| = B$
    \State $\tilde{\mu}^j = \nabla f_{I^j}(\tilde{x}^j)$
    \State $x^j_0 = \tilde{x}^j$
    \State (Option I) Generate $\mathcal{K}^j \sim Geom\left(\frac{B}{B + b}\right)$
    \State (Option II) $\mathcal{K}^j = \frac{B}{b}$
    \For{$k = 1$ to $\mathcal{K}^j$}
        \State Uniformly sample a mini-batch $I^j_k \subset [n]$, s.t. $|I^j_k| = b$
        \State $\mu^j_k = \nabla f_{I^j_k}(x^j_{k-1}) - \nabla f_{I^j_k}(\tilde{x}^j) + \tilde{\mu}^j$
        \State $\tau_k = P(\mu^j_k, \mathbb{M} \oplus \mathbb{M}_\mathcal{T}, \mathbb{M}_\mathcal{H})$
        \State $x^j_k = P(x^j_{k-1} - \eta \tau_k, \mathbb{M}, \mathbb{M}_\mathcal{T})$
    \EndFor
    \State $\tilde{x}^{j+1} = x^j_{\mathcal{K}^j}$
\EndFor
\end{algorithmic}
\end{algorithm}

In contrast to \STO, the inner loop of \SCSG\ computes the gradient over a random set of functions rather than a randomly selected function. \SCSG\ also employs two loops with different batch sizes for gradient calculation, making it more flexible than \STO and \SVRG. This flexibility allows \SCSG\ to serve as a general framework for graph constrained optimization. 

In summary, compared with traditional IHT and StoIHT methods, graph-structured hard thresholding steps mainly differ in their use of Head and Tail Projections. Both proposed new algorithms are much more complex than \STO, introducing distinct variance reduction techniques while maintaining the same sparsification enforcement. It is also important to note that \SVRG\ is a specific scenario of \SCSG, hence, we provide theoretical analysis of \SCSG, which can be easily extended to \SVRG\ case. 

\section{Theoretical Analysis}

\noindent In this section, we present our main theoretical results characterizing the estimation error of parameters $x$.  The proof provides a general framework based on the gradients from \SCSG. Consequently, the main theorem is applicable to our two proposed algorithms, \SVRG and \SCSG. We demonstrate the convergence of these algorithms by bounding the final error using the $\ell^2$ norm of the initial and the optimal distance. 
Additionally, we consider the history of the stochastic process up to iteration $j*\mathcal{K}$ with the notation $I^j_\mathcal{K}$.

Before delving into the main theorem, we present a key lemma that is crucial for the proof.

\begin{lemma}\cite{zhou2019stochastic}
   \label{lemma:x_t}
    If each $f_{\xi_t}(\cdot)$ and $F(x)$ satisfy Assumption \ref{ass:three}, and given head projection model $(c_H, M \oplus M_T, M_H)$ and tail projection model $(c_T, M, M_T)$, then we have the following inequality
\[
\mathbb{E}_{\xi_t} \| (x^t - x^*)_H \| \leq \sqrt{1 - \alpha_0} \mathbb{E}_{\xi_t} \| x^t - x^* \| + \sigma_1,
\]
where
\[
\sigma_1 = \left( \frac{\beta_0}{\alpha_0} + \sqrt{\frac{\alpha_0 \beta_0}{1 - \alpha_0}} \right) \mathbb{E}_{\xi_t} \| \nabla_I f_{\xi_t}(x^*) \|,
\]
\[
H = \text{supp}(P(\nabla f_{\xi_t}(x^t), M \oplus M_T, M_H)),
\]
\[
\alpha_0 = c_H \alpha \tau - \sqrt{\alpha \beta \tau^2 - 2 \alpha \tau + 1}, \quad \beta_0 = (1 + c_H) \tau,
\]
\[
I = \arg \max_{S \in M \oplus M_T \oplus M_H} \mathbb{E}_{\xi_t} \| \nabla_S f_{\xi_t}(x^*) \|,
\]
and $\tau \in (0, 2/\beta)$.
\end{lemma}

With the prepared lemmas and appropriate assumptions in place, we can now present our main theorem. This theorem establishes the convergence properties of our proposed algorithms, under specific conditions. The detailed statement of our main theorem is as follows:

\begin{theorem} (Main Theorem)
\label{theorem1}
Assume that Definition \ref{def:RSC/RSS} and Assumption \ref{ass:three} hold. Under the same setting of Lemma \ref{lemma:x_t}, let $x_0$ be the start point. If we choose a constant learning rate $\eta$ within
\[
\eta \in \left( \frac{2\alpha - \sqrt{4\alpha^2 - 3.75\alpha\beta}}{2\alpha\beta}, \frac{2\alpha +\sqrt{4\alpha^2 - 3.75\alpha\beta}}{2\alpha\beta} \right)
\]
then the solution $\tilde{x}^{j+1}$ of \SCSG\ satisfies
\[
\mathbb{E}_{I^j_{\mathcal{K}}} \| \tilde{x}^{j+1} - x^* \| \leq \left[ \left(\frac{\delta}{1-\lambda}\right)^S + \lambda^j \left( \frac{1-\lambda}{1-\lambda-\delta} \right) \right] \|x^0 - x^* \|
\]
\[
+ \frac{\gamma}{1-\lambda-\delta} \mathbb{E}_{I^j_{\mathcal{K}}} \| \nabla_I f_{I^j}(x^*) \|,
\]
where
\[
\delta = (1 + c_{\mathcal{T}}) \left( \sqrt{\alpha \beta \eta^2 - 2 \alpha \eta + 1} + \sqrt{1 - \alpha_0} \right),
\]
\[
\lambda = (1 + c_{\mathcal{T}}) (2\sqrt{\alpha \beta \eta^2 - 2 \alpha \eta + 1}),
\]
\[
\gamma = (1 + c_{\mathcal{T}})(\frac{\beta_0}{\alpha_0} + \frac{\alpha_0\beta_0}{\sqrt{1-{\alpha_0}^2}} + \eta).
\]
\end{theorem}

Theorem \ref{theorem1} demonstrates that our new algorithm achieves linear convergence with stochastic variance-reduced gradients, even with more stochastic settings in batch and mini-batch.  Each variable defined in the theorem is  strictly less than 1, ensuring that the error decreases as the number of iterations increases. This result aligns with our experimental findings, where more iterations consistently lead to smaller errors. 

From Theorem \ref{theorem1}, we derive the following corollary, which further justifies the convergence of our algorithm and specifies the appropriate range for the learning rate $\eta$.

\begin{corollary}
\label{cor1}
    To ensure convergence of our algorithm, the learning rate $\eta$, which is a constant, should be chosen within the range $( \frac{2\alpha - \sqrt{4\alpha^2 - 3.75\alpha\beta}}{2\alpha\beta}, \frac{2\alpha + \sqrt{4\alpha^2 - 3.75\alpha\beta}}{2\alpha\beta})$. For this range to be valid, the following inequality must hold:
\[
\frac{\delta}{1-\lambda} < 1.
\]
\end{corollary}

Corollary \ref{cor1} is the cornerstone of Theorem \ref{theorem1}. It ensures that the upper bound for the estimation error does not blow up to infinity, and provides a constant value for the finite series. Similarly, it also ensures that the upper bound will decay more after performing more iterations.  Corollary 2.1 also provides a range of $\eta$, which is smaller than the one given by \STO. This way we can find $\eta$ such that the algorithm will always converge. All the proofs are provided in the appendix of the paper.

\section{Experiments}

\subsection{Experimental setup}

\noindent We perform multiple experiments to compare our proposed algorithms with baseline methods. For our experiments, we consider the residual norm of the loss function, $\| \mathbf{A}x^{t+1} - y \|$ as the number of epochs increases. Due to the non-convex nature of the problem, there are several local minima and the algorithm may not approach the global minimum, $x^*$. Additionally, in real-world applications, $x^*$ is often unknown. Therefore, we use the residual norm as a measurement of convergence as opposed to the distance from the final iterate to the target vector, $\|x^{t+1} - x^*\|$. All experiments are tested on a Ubuntu 22.04.4 server with 256 AMD EPYC 9554 64-core processors and with 1.6 TB RAM. All codes are written in Python\footnote{All code is available at https://github.com/Derek-Fox/graph-scsg-iht}. 

\subsection{Synthetic Dataset}
\noindent We first tested our methods on synthetic datasets to determine the optimal parameters. For a fair comparison, we followed the exact settings used in \STO, conducting multiple experiments using a grid graph with a dimension of 256 and unit-weight edges.

\begin{figure*}
    \centering
    \begin{tabular}{ccc}
         \includegraphics[width=0.3\linewidth]{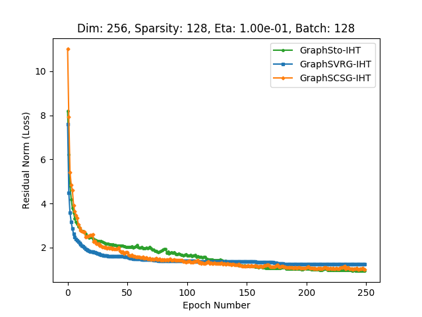}
         & \includegraphics[width=0.3\linewidth]{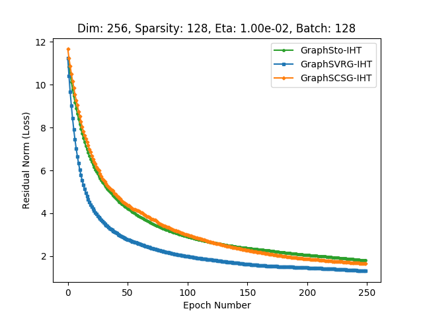}
         & \includegraphics[width=0.3\linewidth]{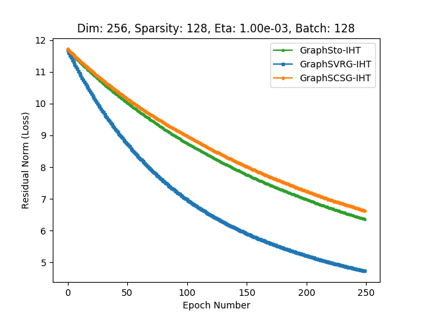}
    \end{tabular}
    \caption{Comparison of methods with different learning rate.}
    \label{fig:1}
\end{figure*}

\begin{figure*}
    \centering
    \begin{tabular}{ccc}
         \includegraphics[width=0.3\linewidth]{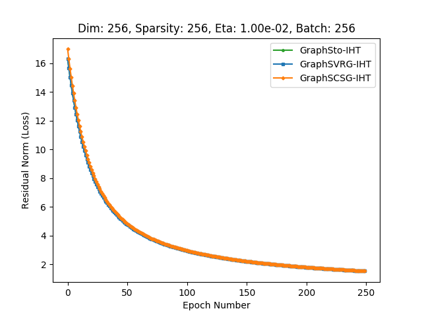} 
         & \includegraphics[width=0.3\linewidth]{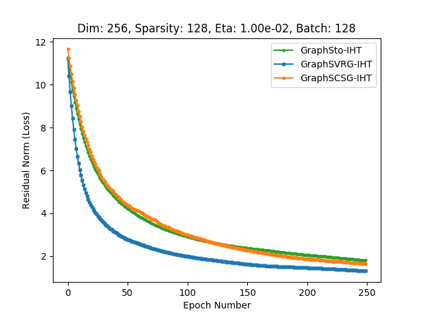} 
         & \includegraphics[width=0.3\linewidth]{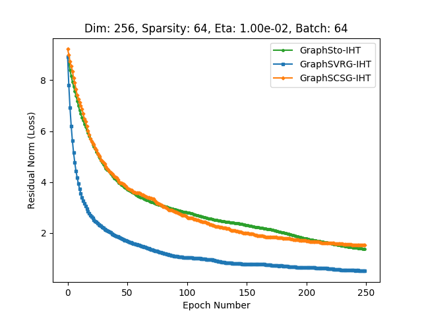}
    \end{tabular}
    \caption{Comparison of methods with different sparsities.}
    \label{fig:2}
\end{figure*}

\begin{figure*}
    \centering
    \begin{tabular}{ccc}
     \includegraphics[width=0.3\linewidth]{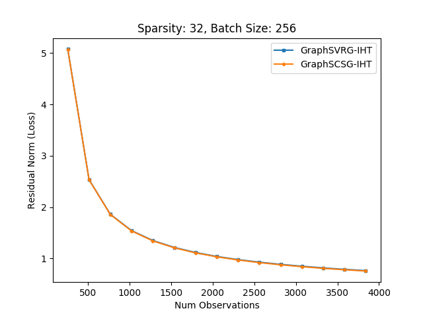} 
    \includegraphics[width=0.3\linewidth]{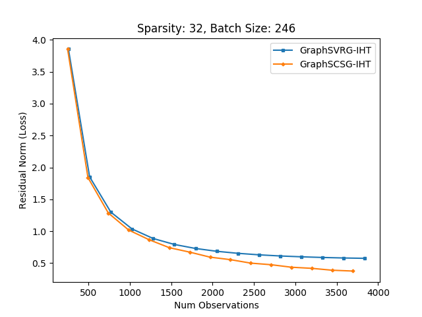} 
    \includegraphics[width=0.3\linewidth]{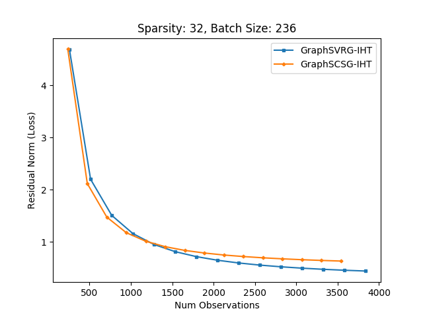} 
    \end{tabular}
    \caption{Number of data points vs. residual loss value with different batch size. }
    \label{fig:3}
\end{figure*}

\noindent \textbf{Choice of $\eta$.} To study the effect of the learning rate on the performance of our algorithms, we varied $\eta$ across \{0.1, 0.01, 0.001\} and tested these rates in various sparsity cases, with sparsity values chosen from  \{256, 128, 64, 32\}. By varying the learning rate, we aimed to understand the convergence behaviour of our algorithms. As shown in Figure \ref{fig:1}, our experiments reveal that, as expected, with a larger learning rate, all methods converge quickly, while with a smaller learning rate, the steps take longer to achieve zero residual loss.  Additionally, both our proposed methods and the baseline method converge stably, and regardless of the learning rate setting, \SVRG\ consistently performs the best. 

\noindent \textbf{Choice of sparsity.} Studying the setting of sparsity, $s$, is crucial for understanding the performance of our algorithms in graph sparsification optimization. To examine the effect of sparsity, we compared our methods, \SVRG\ and \SCSG\, against the baseline algorithm  \STO. 
Using the experimental settings from \STO\, we employed a grid graph of dimension 256, fixed the learning rate $\eta = 0.01$ and set the batch size equal to the sparsity parameter $B = s$. We varied the sparsity parameter $s$ to observe the behavior of all algorithms, as shown in Figure \ref{fig:2}. Figure \ref{fig:2} demonstrates that as the sparsity parameter $s$ decreases, \SVRG\ outperforms the other algorithms in minimizing the residual norm $\| \mathbf{A}x^{t+1} - y \|$ over epochs. Another interesting finding is that we observed that \STO\ and \SCSG\ display almost identical behavior in this parameter setting.

\begin{table*}[t]
\caption{Identified genes from breast cancer dataset}
\label{table:genes}
\centering
\begin{tabular}{lcl}
\toprule
    Method & Number & Genes \\
\midrule
IHT      & 2 & NAT1 TOP2A \\
\SIHT    & 2 & NAT1 TOP2A \\
\STO     & 6  & AR ATM BRCA2 CCND2 CDKN1A TOP2A \\
\textbf{\SVRG}    & 6  & AR ATM BRCA2 CCND2 CDKN1A TOP2A \\
\textbf{\SCSG}    & 10 & AR ATM BRCA1 BRCA2 CCND2 CCND3 CDKN1A CHEK2 FBP1 TOP2A \\
\bottomrule
\end{tabular}
\end{table*}

\noindent \textbf{Choice of batch size.}  After exploring the choice of $\eta$ and $s$, we varied the batch size $B$ on a grid graph with a dimension of 256 to demonstrate the advantages of \SCSG. Here we fixed $\eta = 0.01$, $s = 32$. For fair comparison, we considered the number of data points instead of the number of epochs to estimate the run time of the algorithms, which is a common practice in optimization.

Since gradient calculations are computationally expensive, using fewer data points would result in faster run times. Figure \ref{fig:3} shows three different scenarios with varying batch sizes $B$. When $B$ equals to the dimension, \SCSG\ degrades to \SVRG\, resulting in similar performance. With smaller $B$ (meaning fewer data points are used in the full gradient calculation), \SCSG\ consistently outperforms \SVRG. However, as $B$ continues to decrease, \SCSG\ initially outperforms \SVRG\, but was eventually surpassed by \SVRG\ slightly. This occurs because smaller batch size increase gradient variance, making it harder for \SCSG\ to maintain its initial advantage. Furthermore, the interaction between batch size and mini-batch size becomes more complex, ultimately affecting the final performance.

Therefore, we also conducted numerous experiments to study the effects of mini-batch size in conjunction with batch size. Additionally, to better understand the graph-structure patterns, we have explored how varying the number of connected components in subgraphs impacts final convergence performance. Additional results are provided in the Appendix.

\subsection{Real-world Dataset}

\noindent To test our algorithms on a real-world dataset, we use a large breast cancer dataset \cite{van2002gene}. This dataset contains 295 training samples with 8,141 genes dimensions, including 78 positive (metastatic) and 217 negative (non-metastatic) samples. Following the experimental setting in \cite{zhou2019stochastic}, we use a Protein-Protein Interaction network with 637 pathways from \cite{jacob2009group}, the dataset is folded into 5 subfolds, and 20 trials are conducted. 

Table \ref{table:genes} shows the results from the gene identification task on the breast cancer dataset. \SCSG\ identifies 40\% of the 25 genes highly correlated with breast cancer, as gathered by \cite{zhou2019stochastic}. \STO\ and \SVRG\ both identify 24\% of these genes, consistent with the findings in \cite{zhou2019stochastic}. All the graph-structured methods greatly outperform the IHT and \SIHT\ methods which only identify 8\% of the cancer-related genes. These results demonstrate the promise of variance-reduction techniques for stochastic gradient descent in the setting of graph sparsity optimization.

In summary, \SVRG\ is more stable while \SCSG\ shows excellent performance under appropriate settings due to the introduction of two additional batch size. This allows for greater flexibility and control in the gradient estimation process. Overall, our two new algorithms demonstrate both efficiency and effectiveness by incorporating variance reduction techniques in graph sparsity problems, making them suitable for a wide range of applications.

\section{Conclusion}
\noindent We have proposed two algorithms to utilize variance reduction techniques in the setting of graph sparse optimization. The proposed algorithms can significantly improve the stability and efficiency in machine learning and other applications. Theoretically, we provide a general proof framework showing linear convergence. Empirically, our algorithms are competitive in minimizing the objective loss function compared to their predecessors in various experimental settings with a synthetic dataset. Additionally, testing on a large-scale medical dataset demonstrated superior performance in identifying cancer-related genes. Future work should include testing our algorithms on more larger real-world datasets. By employing graph-structured hard thresholding, we can uncover more underlying subgraphs and related patterns, with significant implications in fields such as medical research and social network analysis. This approach can enhance our understanding of complex data structures and lead to more effective solutions.

\section{Acknowledgement}
\noindent This work was completed at the University of North Carolina at Greensboro, funded by NSF REU program 2349369.

\bibliographystyle{plain}
\bibliography{reference}

\begin{thebibliography}{10}

\bibitem{aksoylar2017connected}
Cem Aksoylar, Lorenzo Orecchia, and Venkatesh Saligrama.
\newblock Connected subgraph detection with mirror descent on sdps.
\newblock In {\em International Conference on Machine Learning}, pages 51--59. PMLR, 2017.

\bibitem{arias2011detection}
Ery Arias-Castro, Emmanuel~J Candes, and Arnaud Durand.
\newblock Detection of an anomalous cluster in a network.
\newblock {\em The Annals of Statistics}, pages 278--304, 2011.

\bibitem{bach2012structured}
Francis Bach, Rodolphe Jenatton, Julien Mairal, and Guillaume Obozinski.
\newblock Structured sparsity through convex optimization.
\newblock {\em Statistical Science}, 27(4):450--468, 2012.

\bibitem{bach2012optimization}
Francis Bach, Rodolphe Jenatton, Julien Mairal, Guillaume Obozinski, et~al.
\newblock Optimization with sparsity-inducing penalties.
\newblock {\em Foundations and Trends{\textregistered} in Machine Learning}, 4(1):1--106, 2012.

\bibitem{bahmani2013greedy}
Sohail Bahmani, Bhiksha Raj, and Petros~T Boufounos.
\newblock Greedy sparsity-constrained optimization.
\newblock {\em Journal of Machine Learning Research}, 14(Mar):807--841, 2013.

\bibitem{baraniuk2010model}
Richard~G Baraniuk, Volkan Cevher, Marco~F Duarte, and Chinmay Hegde.
\newblock Model-based compressive sensing.
\newblock {\em IEEE Transactions on information theory}, 56(4):1982--2001, 2010.

\bibitem{blumensath2009iterative}
Thomas Blumensath and Mike~E Davies.
\newblock Iterative hard thresholding for compressed sensing.
\newblock {\em Applied and computational harmonic analysis}, 27(3):265--274, 2009.

\bibitem{chen2001atomic}
Scott~Shaobing Chen, David~L Donoho, and Michael~A Saunders.
\newblock Atomic decomposition by basis pursuit.
\newblock {\em SIAM review}, 43(1):129--159, 2001.

\bibitem{defazio2014saga}
Aaron Defazio, Francis Bach, and Simon Lacoste-Julien.
\newblock Saga: A fast incremental gradient method with support for non-strongly convex composite objectives.
\newblock In {\em Advances in neural information processing systems}, pages 1646--1654, 2014.

\bibitem{foucart2011hard}
Simon Foucart.
\newblock Hard thresholding pursuit: an algorithm for compressive sensing.
\newblock {\em SIAM Journal on Numerical Analysis}, 49(6):2543--2563, 2011.

\bibitem{hegde2014fast}
Chinmay Hegde, Piotr Indyk, and Ludwig Schmidt.
\newblock A fast approximation algorithm for tree-sparse recovery.
\newblock In {\em 2014 IEEE International Symposium on Information Theory}, pages 1842--1846. IEEE, 2014.

\bibitem{hegde2015approximation}
Chinmay Hegde, Piotr Indyk, and Ludwig Schmidt.
\newblock Approximation algorithms for model-based compressive sensing.
\newblock {\em IEEE Transactions on Information Theory}, 61(9):5129--5147, 2015.

\bibitem{hegde2015nearly}
Chinmay Hegde, Piotr Indyk, and Ludwig Schmidt.
\newblock A nearly-linear time framework for graph-structured sparsity.
\newblock In {\em International Conference on Machine Learning}, pages 928--937. PMLR, 2015.

\bibitem{hegde2016fast}
Chinmay Hegde, Piotr Indyk, and Ludwig Schmidt.
\newblock Fast recovery from a union of subspaces.
\newblock {\em Advances in Neural Information Processing Systems}, 29, 2016.

\bibitem{jacob2009group}
Laurent Jacob, Guillaume Obozinski, and Jean-Philippe Vert.
\newblock Group lasso with overlap and graph lasso.
\newblock In {\em Proceedings of the 26th annual international conference on machine learning}, pages 433--440, 2009.

\bibitem{jain2014iterative}
Prateek Jain, Ambuj Tewari, and Purushottam Kar.
\newblock On iterative hard thresholding methods for high-dimensional m-estimation.
\newblock In {\em Advances in Neural Information Processing Systems}, pages 685--693, 2014.

\bibitem{johnson2013accelerating}
Rie Johnson and Tong Zhang.
\newblock Accelerating stochastic gradient descent using predictive variance reduction.
\newblock In {\em Advances in neural information processing systems}, pages 315--323, 2013.

\bibitem{lei2017less}
Lihua Lei and Michael Jordan.
\newblock Less than a single pass: Stochastically controlled stochastic gradient.
\newblock In {\em Artificial Intelligence and Statistics}, pages 148--156, 2017.

\bibitem{li2016stochastic}
Xingguo Li, Tuo Zhao, Raman Arora, Han Liu, and Jarvis Haupt.
\newblock Stochastic variance reduced optimization for nonconvex sparse learning.
\newblock In {\em International Conference on Machine Learning}, pages 917--925, 2016.

\bibitem{liang2020effective}
Guannan Liang, Qianqian Tong, Chunjiang Zhu, and Jinbo Bi.
\newblock An effective hard thresholding method based on stochastic variance reduction for nonconvex sparse learning.
\newblock In {\em Proceedings of the AAAI Conference on Artificial Intelligence}, volume 34(02), pages 1585--1592, 2020.

\bibitem{nguyen2017linear}
Nam Nguyen, Deanna Needell, and Tina Woolf.
\newblock Linear convergence of stochastic iterative greedy algorithms with sparse constraints.
\newblock {\em IEEE Transactions on Information Theory}, 63(11):6869--6895, 2017.

\bibitem{qian2014connected}
Jing Qian, Venkatesh Saligrama, and Yuting Chen.
\newblock Connected sub-graph detection.
\newblock In {\em Artificial Intelligence and Statistics}, pages 796--804. PMLR, 2014.

\bibitem{rozenshtein2014event}
Polina Rozenshtein, Aris Anagnostopoulos, Aristides Gionis, and Nikolaj Tatti.
\newblock Event detection in activity networks.
\newblock In {\em Proceedings of the 20th ACM SIGKDD international conference on Knowledge discovery and data mining}, pages 1176--1185, 2014.

\bibitem{tibshirani1996regression}
Robert Tibshirani.
\newblock Regression shrinkage and selection via the lasso.
\newblock {\em Journal of the Royal Statistical Society: Series B (Methodological)}, 58(1):267--288, 1996.

\bibitem{turlach2005simultaneous}
Berwin~A Turlach, William~N Venables, and Stephen~J Wright.
\newblock Simultaneous variable selection.
\newblock {\em Technometrics}, 47(3):349--363, 2005.

\bibitem{van2008high}
Sara~A Van~de Geer et~al.
\newblock High-dimensional generalized linear models and the lasso.
\newblock {\em The Annals of Statistics}, 36(2):614--645, 2008.

\bibitem{van2002gene}
Marc~J Van De~Vijver, Yudong~D He, Laura~J Van't~Veer, Hongyue Dai, Augustinus~AM Hart, Dorien~W Voskuil, George~J Schreiber, Johannes~L Peterse, Chris Roberts, Matthew~J Marton, et~al.
\newblock A gene-expression signature as a predictor of survival in breast cancer.
\newblock {\em New England Journal of Medicine}, 347(25):1999--2009, 2002.

\bibitem{yuan2006model}
Ming Yuan and Yi~Lin.
\newblock Model selection and estimation in regression with grouped variables.
\newblock {\em Journal of the Royal Statistical Society Series B: Statistical Methodology}, 68(1):49--67, 2006.

\bibitem{yuan2014gradient}
Xiaotong Yuan, Ping Li, and Tong Zhang.
\newblock Gradient hard thresholding pursuit for sparsity-constrained optimization.
\newblock In {\em International Conference on Machine Learning}, pages 127--135, 2014.

\bibitem{zhou2019stochastic}
Baojian Zhou, Feng Chen, and Yiming Ying.
\newblock Stochastic iterative hard thresholding for graph-structured sparsity optimization.
\newblock In {\em International Conference on Machine Learning}, pages 7563--7573. PMLR, 2019.

\bibitem{zhou2018efficient}
Pan Zhou, Xiaotong Yuan, and Jiashi Feng.
\newblock Efficient stochastic gradient hard thresholding.
\newblock In {\em Advances in Neural Information Processing Systems}, pages 1988--1997, 2018.

\end{thebibliography}

\onecolumn
\appendices
\section{Parameters Table}

\begin{center}\label{table:parameters}
Table: Constraints for Parameters
\end{center}
\begin{center}
\begin{tabular}{|c|c|c|} 
  \hline
  \textbf{Notation} & \textbf{Name} & \textbf{Constraint} \\ 
  \hline
  $\eta$ & Step size & Fixed \\ 
  \hline
  F() & Function & Minimized \\ 
  \hline
  $\mathbb{M}$ & Set of Supports &\\
  \hline
  $\mathbb{M_\mathcal{H}}$ & Set of Supports & \\
  \hline
  $\mathbb{M_\mathcal{T}}$ & Set of Supports & \\
  \hline
  B & Big batch & \\
  \hline
  $I^j$ & Big batch sample & $I^j \subset [n]$ and $|I^j| = B$ \\
  \hline
  b & Small batch & Subset of B\\
  \hline
  $I^j_k$ & Small batch sample & $I^j_k \subset [n]$ and $|I^j_k| = b$\\
  \hline
  $\mathcal{J}$ & Number of outer loops & \\
  \hline
  j & Outer loop index &\\
  \hline
  $\mathcal{K}$ & Number of inner loops & \\
  \hline
  k & Inner loop index &\\
  \hline
  $\tilde{x} ^ 1$ & Starting Position & $supp(\tilde{x} ^ 1)\in \mathbb{M}$ \\
  \hline
  s & Number of non-zero entries & Sparsity Constraint\\
  \hline
  $P(x, \mathbb{M}, \mathbb{M}_\mathcal{H})$ & Head Projection & \\
  \hline
  $P(x, \mathbb{M}, \mathbb{M}_\mathcal{T})$ & Tail Projection & \\
  \hline
  $\tilde{v}^j$ & SVRG Gradient &\\
  \hline
  $\tilde{\mu}^j$ & SCSG Gradient &\\
  \hline
  $\tilde{x}^j$ & Current Position &\\
  \hline
  $v^j_k$ & Reduced Variance Gradient &\\
  \hline
  $\mu^j_k$ & Reduced Variance Gradient &\\
  \hline
  $\tau_k$ & Sparsified Gradient &\\
  \hline
  $g$ & \small Number of Connected Components &\\
  \hline
\end{tabular}
\end{center}

\section{Proof}

\noindent \textbf{Lemma 4 \cite{nguyen2017linear}.} \label{lemma:4} Let $\xi_t$ be a discrete random variable defined on $[n]$ and its probability mass function is defined as $\Pr(\xi_t = i) := \frac{1}{n}$, which means the probability of $\xi_t$ selecting the $i$-th block at time $t$. For any fixed sparse vectors $x$, $y$ and $0 < \tau < \frac{2}{\beta}$, we have
\[
\mathbb{E}_{\xi_t} \| x - y - \tau (\nabla_\Omega f_{\xi_t}(x) - \nabla_\Omega f_{\xi_t}(y)) \| \leq \sqrt{\alpha \beta \tau^2 - 2\alpha \tau + 1} \|x - y\|,
\]
where $\Omega$ is such that $\|x\|_0 \cup \|y\|_0 \subseteq \Omega$ and $\Omega \in \mathcal{M}$.

\begin{proof}
 We first try to obtain an upper bound for $\mathbb{E}_{\xi_t} \|x - y - \tau (\nabla_\Omega f_{\xi_t}(x) - \nabla_\Omega f_{\xi_t}(y))\|^2$ as follows:
\begin{align*}
\mathbb{E}_{\xi_t} \| x - y - \tau (\nabla_\Omega f_{\xi_t}(x) - \nabla_\Omega f_{\xi_t}(y)) \|^2 &= \|x - y\|^2 - 2\tau \mathbb{E}_{\xi_t} \langle x - y, \nabla_\Omega f_{\xi_t}(x) - \nabla_\Omega f_{\xi_t}(y) \rangle \\
&\quad + \tau^2 \mathbb{E}_{\xi_t} \|\nabla_\Omega f_{\xi_t}(x) - \nabla_\Omega f_{\xi_t}(y)\|^2 \\
&= \|x - y\|^2 - 2\tau \langle x - y, \mathbb{E}_{\xi_t} [\nabla f_{\xi_t}(x) - \nabla f_{\xi_t}(y)] \rangle \\
&\quad + \tau^2 \mathbb{E}_{\xi_t} \|\nabla_\Omega f_{\xi_t}(x) - \nabla_\Omega f_{\xi_t}(y)\|^2 \\
&\leq \|x - y\|^2 - 2\tau \langle x - y, \nabla F(x) - \nabla F(y) \rangle \\
&\quad + \tau^2 \beta \langle x - y, \nabla F(x) - \nabla F(y) \rangle \\
&= \|x - y\|^2 + (\tau^2 \beta - 2\tau) \langle x - y, F(x) - F(y) \rangle \\
&\leq (\alpha \beta \tau^2 - 2\alpha \tau + 1)\|x - y\|^2,
\end{align*}
where the second equality uses the fact that $\|x\|_0 \cup \|y\|_0 \subseteq \Omega$, the first inequality follows from Lemma 2, the third equality is obtained by using the fact that $\mathbb{E}_{\xi_t}[\nabla f_{\xi_t}(x) - \nabla f_{\xi_t}(y)] = \nabla F(x) - \nabla F(y)$, and the last inequality is due to the restricted strong convexity of $F(x)$ on $\mathcal{M(M)}$. We complete the proof by taking the square root of both sides and using the fact: for any random variable $X$, we have $(\mathbb{E}[X])^2 \leq \mathbb{E}[X^2]$.
\end{proof}

\noindent\textbf{Proof of Theorem 2.}
\begin{proof}
\setcounter{equation}{0}
\begin{align}
\mathbb{E}_{I^j_\mathcal{K}} \| \Tilde{x}^{j+1} - x^* \|
&= \mathbb{E}_{I^j_\mathcal{K} | I^j_{\mathcal{K}-1}} \| P(x^j_{\mathcal{K}-1} - \eta \upsilon_\mathcal{K}, \mathbb{M}, \mathbb{M}_T) - x^* \| \\
\begin{split}
&\leq \mathbb{E}_{I^j_\mathcal{K} | I^j_{\mathcal{K}-1}} \| P(x^j_{\mathcal{K}-1} - \eta \upsilon_\mathcal{K}, \mathbb{M}, \mathbb{M}_T) - (x^j_{\mathcal{K}-1} - \eta \upsilon_\mathcal{K}) \| \\
&\quad + \mathbb{E}_{I^j_\mathcal{K} | I^j_{\mathcal{K}-1}} \| (x^j_{\mathcal{K}-1} - \eta \upsilon_{\mathcal{K}}) - x^* \|
\end{split} \\
&\leq (1 + c_{\mathcal{T}}) \mathbb{E}_{I^j_\mathcal{K} | I^j_{\mathcal{K}-1}} \|x^j_{\mathcal{K}-1} - \eta \upsilon_{\mathcal{K}} - x^* \| \\
&= (1 + c_{\mathcal{T}}) \mathbb{E}_{I^j_\mathcal{K} | I^j_{\mathcal{K}-1}} \|r^{\mathcal{K}} - \eta \upsilon_{\mathcal{K}} \| \\
\begin{split}
&= (1 + c_{\mathcal{T}}) \mathbb{E}_{I^j_\mathcal{K} | I^j_{\mathcal{K}-1}} \|r^{\mathcal{K}} - \eta (\nabla_H f_{I^j_{\mathcal{K}}}(x^j_{\mathcal{K}-1}) \\
&\quad - \nabla_H f_{I^j_{\mathcal{K}}}(\Tilde{x}^j) + \nabla_H f_{I^j}(\Tilde{x}^j) ) \| \\
\end{split}
\end{align}

By the definition of $\Tilde{x}^{j+1}$, Eq. (1) holds. Eq. (2) holds by the Triangle Inequality. Eq. (3) holds by Assumption \ref{ass:tail}. Eq. (4) holds by defining $r^{\mathcal{K}} = x^j_{\mathcal{K}-1}  - x^*$. Eq. (5) holds by the definition of $\upsilon_{\mathcal{K}}$. Now, we will focus on the expectation only. Also, define $m^{\mathcal{K}} = \Tilde{x}^{j} - x^*$.

\begin{align}
\Rightarrow& \ \mathbb{E}_{I^j_\mathcal{K} | I^j_{\mathcal{K}-1}} \| r^\mathcal{K} - \eta (\nabla_H f_{I^j_{\mathcal{K}}}(x^j_{\mathcal{K}-1}) - \nabla_H f_{I^j_{\mathcal{K}}}(\Tilde{x}^j) + \nabla_H f_{I^j}(\Tilde{x}^j) ) \| \notag \\
\begin{split}
&\quad\leq \mathbb{E}_{I^j_\mathcal{K} | I^j_{\mathcal{K}-1}} [\| r^\mathcal{K}_H - \eta (\nabla_H f_{I^j_{\mathcal{K}}}(x^j_{\mathcal{K}-1}) - \nabla_H f_{I^j_{\mathcal{K}}}(\Tilde{x}^j) + \nabla_H f_{I^j}(\Tilde{x}^j) ) \\
&\quad - (\nabla_H f_{I^j_{\mathcal{K}}}(x^*) - \nabla_H f_{I^j_{\mathcal{K}}}(x^*) + \nabla_H f_{I^j}(x^*) ) \| \\
&\quad + \eta \| \nabla_H f_{I^j}(x^*) \| + \|r^\mathcal{K}_{H^c} \| ]
\end{split}\\
\begin{split}
&\quad \leq \mathbb{E}_{I^j_\mathcal{K} | I^j_{\mathcal{K}-1}} [\| r^\mathcal{K}_{H \cup \Omega} - \eta (\nabla_{H \cup \Omega} f_{I^j_{\mathcal{K}}}(x^j_{\mathcal{K}-1}) - \nabla_{H \cup \Omega} f_{I^j_{\mathcal{K}}}(\Tilde{x}^j) \\
&\quad + \nabla_{H \cup \Omega} f_{I^j}(\Tilde{x}^j) ) - (\nabla_{H \cup \Omega} f_{I^j_{\mathcal{K}}}(x^*) - \nabla_{H \cup \Omega} f_{I^j_{\mathcal{K}}}(x^*) + \nabla_{H \cup \Omega} f_{I^j}(x^*) ) \| \\
&\quad+ \eta \| \nabla_H f_{I^j}(x^*) \| + \|r^\mathcal{K}_{H^c} \| ]
\end{split}\\
\begin{split}
&\quad= \mathbb{E}_{I^j_\mathcal{K} | I^j_{\mathcal{K}-1}} [\| r^\mathcal{K}_{H \cup \Omega} -m^\mathcal{K}_{H \cup \Omega} + m^\mathcal{K}_{H \cup \Omega} - \eta (\nabla_{H \cup \Omega} f_{I^j_{\mathcal{K}}}(x^j_{\mathcal{K}-1}) \\
&\quad - \nabla_{H \cup \Omega} f_{I^j_{\mathcal{K}}}(\Tilde{x}^j) + \nabla_{H \cup \Omega} f_{I^j}(\Tilde{x}^j) ) - (\nabla_{H \cup \Omega} f_{I^j_{\mathcal{K}}}(x^*) - \nabla_{H \cup \Omega} f_{I^j_{\mathcal{K}}}(x^*) \\
&\quad + \nabla_{H \cup \Omega} f_{I^j}(x^*) ) \| + \eta \| \nabla_H f_{I^j}(x^*) \| + \|r^\mathcal{K}_{H^c} \| ]
\end{split}\\
\begin{split}
&\quad \leq \mathbb{E}_{I^j_\mathcal{K} | I^j_{\mathcal{K}-1}} [\| r^\mathcal{K}_{H \cup \Omega} - \eta (\nabla_{H \cup \Omega} f_{I^j_{\mathcal{K}}}(x^j_{\mathcal{K}-1}) - \nabla_{H \cup \Omega} f_{I^j_{\mathcal{K}}}(\Tilde{x}^j) \\
&\quad + \nabla_{H \cup \Omega} f_{I^j}(\Tilde{x}^j) ) - (\nabla_{H \cup \Omega} f_{I^j_{\mathcal{K}}}(x^*) - \nabla_{H \cup \Omega} f_{I^j_{\mathcal{K}}}(x^*) + \nabla_{H \cup \Omega} f_{I^j}(x^*) ) \| \\
&\quad+ \eta \| \nabla_H f_{I^j}(x^*) \| + \|r^\mathcal{K}_{H^c} \| ]\\
&\quad \leq \sqrt{\alpha \beta \eta^2 - 2 \alpha \eta + 1} \mathbb{E}_{I^j_{\mathcal{K}-1}} \|r_{\mathcal{K}}\| + \sqrt{\alpha \beta \eta^2 - 2 \alpha \eta + 1} \mathbb{E}_{I^j_{\mathcal{K}-1}} \|-m_{\mathcal{K}}\| \\
&\quad + \sqrt{\alpha \beta \eta^2 - 2 \alpha \eta + 1} \mathbb{E}_{I^j_{\mathcal{K}-1}} \|m_{\mathcal{K}}\| + \eta \mathbb{E}_{I^j_{\mathcal{K}}} \|\nabla_I f_{I^j}(x^*)\| + \mathbb{E}_{{I^j_{\mathcal{K}}}|{I^j_{\mathcal{K}-1}}} \|r^{\mathcal{K}}_{H^{c}}\|
\end{split}\\
\begin{split}
&\quad = \sqrt{\alpha \beta \eta^2 - 2 \alpha \eta + 1} \mathbb{E}_{I^j_{\mathcal{K}-1}} \|r_{\mathcal{K}}\| + 2\sqrt{\alpha \beta \eta^2 - 2 \alpha \eta + 1} \mathbb{E}_{I^j_{\mathcal{K}-1}} \|m_{\mathcal{K}}\| \\
&\quad + \eta \mathbb{E}_{I^j_{\mathcal{K}}} \|\nabla_I f_{I^j}(x^*)\| + \mathbb{E}_{{I^j_{\mathcal{K}}}|{I^j_{\mathcal{K}-1}}} \|r^{\mathcal{K}}_{H^{c}}\|
\end{split} \\
\begin{split}
&\quad \leq (\sqrt{\alpha \beta \eta^2 - 2 \alpha \eta + 1} + \sqrt{1-{\alpha_0}^2}) \mathbb{E}_{I^j_{\mathcal{K}}} \|x^j_{\mathcal{K}^j-1}-x^*\| \\
&\quad + 2\sqrt{\alpha \beta \eta^2 - 2 \alpha \eta + 1} \mathbb{E}_{I^j_{\mathcal{K}-1}} \|\Tilde{x}^{j}-x^*\| \\
&\quad + \left(\frac{\beta_0}{\alpha_0} + \frac{\alpha_0\beta_0}{\sqrt{1-{\alpha_0}^2}} + \eta\right)\mathbb{E}_{I^j_{\mathcal{K}}} \|\nabla_I f_{I^j}(x^*)\|
\end{split}
\end{align}

Eq. (6) and Eq. (7) hold by the Triangle Inequality. Eq. (8) holds by adding and subtracting $m_{\mathcal{K}}$. Eq. (9) also holds by the Triangle Inequality. Eq. (10) holds by \ref{lemma:4}. Eq. (11) holds by the fact that $\| m_{\mathcal{K}}\| =  \| -m_{\mathcal{K}}\|$. Eq. (12) holds by \ref{lemma:x_t}. We therefore get the following inequality that we will use to find our final equation.
\begin{align}
    \begin{split}
        \bb{E}_{I^j_\K} \| \Tilde{x}^{j+1} - x^* \| &\leq \delta \bb{E}_{I^j_{\mathcal{K}}} \|x^j_{\mathcal{K}^j-1}-x^*\| + \lambda\bb{E}_{I^j_{\mathcal{K}-1}} \|\Tilde{x}^{j}-x^*\|\\
        &\quad+  (1 + c_{\mathcal{T}})(\frac{\beta_0}{\alpha_0} \quad + \frac{\alpha_0\beta_0}{\sqrt{1-{\alpha_0}^2}} + \eta)\bb{E}_{I^j_{\mathcal{K}}} \|\nabla_I f_{I^j}(x^*)\|
    \end{split}
\end{align}
where $\delta = (1 + c_{\mathcal{T}})(\sqrt{\alpha \beta \eta^2 - 2 \alpha \eta + 1} + \sqrt{1-{\alpha_0}^2})$ and
$\lambda = 2(1 + c_{\mathcal{T}})(\frac{\delta}{1 + c_{\mathcal{T}}} - \sqrt{1-{\alpha_0}^2})$
Since we have two different vectors measured from the target vector $x^*$, we will use recursion twice in Eq. (13) to find the formula.

First, we apply recursion with respect to the outer loop, j, by finding $\Tilde{x}^{j}$ using the above inequality. Once we find it, we plug it back to understand the behavior and do this process j times.
\begin{align}
\begin{split}
\bb{E}_{I^j_\K} \| \Tilde{x}^{j} - x^* \| &\leq \delta \bb{E}_{I^j_{\mathcal{K}}} \|x^j_{\mathcal{K}^j-1}-x^*\| + \lambda\bb{E}_{I^j_{\mathcal{K}-1}} \|\Tilde{x}^{j-1}-x^*\|\\
&\quad+ (1 + c_{\mathcal{T}})(\frac{\beta_0}{\alpha_0} \quad + \frac{\alpha_0\beta_0}{\sqrt{1-{\alpha_0}^2}} + \eta)\bb{E}_{I^j_{\mathcal{K}}} \|\nabla_I f_{I^j}(x^*)\|
\end{split}   
\end{align}
Plugging Eq. (14) in Eq. (13), we get:
\begin{align}
\begin{split}
\bb{E}_{I^j_\K} \| \Tilde{x}^{j+1} - x^* \| &\leq \delta \bb{E}_{I^j_{\mathcal{K}}} \|x^j_{\mathcal{K}^j-1}-x^*\| + \lambda (\delta \bb{E}_{I^j_{\mathcal{K}}} \|x^j_{\mathcal{K}^j-1}-x^*\| + \lambda\bb{E}_{I^j_{\mathcal{K}-2}} \|\Tilde{x}^{j-1}-x^*\| \\
&\quad + (1 + c_{\mathcal{T}})(\frac{\beta_0}{\alpha_0} + \frac{\alpha_0\beta_0}{\sqrt{1-{\alpha_0}^2}} + \eta)\bb{E}_{I^j_{\mathcal{K}}} \|\nabla_I f_{I^j}(x^*)\|
\end{split}
\end{align}
After doing it j times, we get:
\begin{align}
\begin{split}
\bb{E}_{I^j_\K} \| \Tilde{x}^{j+1} - x^* \| &\leq ( \sum_{i=0}^j \lambda^i) \delta  \bb{E}_{I^j_{\K-1}} \| x^j_{\K^j-1} - x^* \| \ + \ \lambda^j \|x^0 - x^* \| \\
&\quad + \ ( \sum_{i=0}^j \lambda^i) (1 + c_{\mathcal{T}})(\frac{\beta_0}{\alpha_0} + \frac{\alpha_0\beta_0}{\sqrt{1-{\alpha_0}^2}} + \eta) \bb{E}_{I^j_\K} \| \nabla_I f_{I^j}(x^*) \|
\end{split}
\end{align}
By geomteric series (since $\lambda < 1$), and by defining $\gamma = (1 + c_{\mathcal{T}})(\frac{\beta_0}{\alpha_0} + \frac{\alpha_0\beta_0}{\sqrt{1-{\alpha_0}^2}} + \eta)$, we get: 
\begin{align}
\begin{split}
\bb{E}_{I^j_\K} \| \Tilde{x}^{j+1} - x^* \| &\leq (\frac{1}{1-\lambda}) \delta  \bb{E}_{I^j_{\K-1}} \| x^j_{\K^j-1} - x^* \| \ + \ \lambda^j \|x^0 - x^* \| \\
&\quad + \ (\frac{1}{1-\lambda})\gamma\bb{E}_{I^j_\K} \| \nabla_I f_{I^j}(x^*) \|
\end{split}
\end{align}
Once we did the recursion with respect to the outer loop only, we do recursion with respect the inner loop and outer loop to find the final formula. This means that we will do this process S times, where $S = \frac{\mathcal{B*T}}{b}$.
\begin{align}
\begin{split}
\bb{E}_{I^j_{\K-1}} \| x^j_{\K^j-1} - x^* \| &\leq (\frac{1}{1-\lambda}) \delta  \bb{E}_{I^j_{\K-2}} \| x^j_{\K^j-2} - x^* \| \ + \ \lambda^j \|x^0 - x^* \| \\
&\quad + \ (\frac{1}{1-\lambda})\gamma\bb{E}_{I^j_\K} \| \nabla_I f_{I^j}(x^*) \|
\end{split}
\end{align}
We then plug Eq. (18) into Eq. (17), so we get Eq. (19):
\begin{align}
\begin{split}
\bb{E}_{I^j_\K} \| \Tilde{x}^{j+1} - x^* \| &\leq (\frac{\delta}{1-\lambda})[(\frac{\delta}{1-\lambda})\bb{E}_{I^j_{\K-2}} \| x^j_{\K^j-2} - x^* \| \ \\
&\quad + \ \lambda^j \|x^0 - x^* \| + (\frac{\gamma}{1-\lambda})\bb{E}_{I^j_\K} \| \nabla_I f_{I^j}(x^*) \| ] \\
&\quad + \lambda^j \|x^0 - x^* \| + (\frac{\gamma}{1-\lambda})\bb{E}_{I^j_\K} \| \nabla_I f_{I^j}(x^*) \|
\end{split}
\end{align}
After doing this step S times, we get the following inequality:
\begin{align}
\begin{split}
\bb{E}_{I^j_\K} \| \Tilde{x}^{j+1} - x^* \| &\leq
(\frac{\delta}{1-\lambda})^S\|x^0 - x^*\|+ \lambda^j(\sum_{n=0}^S(\frac{\delta}{1-\lambda})^n)\|x^0 - x^* \|\\
&\quad + \frac{\gamma}{1-\lambda}(\sum_{n=1}^S (\frac{\delta}{1-\lambda})^n) \bb{E}_{I^j_\K} \| \nabla_I f_{I^j}(x^*) \|
\end{split}
\end{align}
We finally get this formula that guarantees the convergence of the algorithm, since $\delta$ and $\lambda$ are both less than 1.
\[
\bb{E}_{I^j_\K} \| \Tilde{x}^{j+1} - x^* \| \leq [(\frac{\delta}{1-\lambda})^S + \lambda^j (\sum_{n=0}^S(\frac{\delta}{1-\lambda})^n)] \|x^0 - x^* \| + \frac{\gamma}{1-\lambda}(\sum_{n=0}^S (\frac{\delta}{1-\lambda})^n) \bb{E}_{I^j_\K} \| \nabla_I f_{I^j}(x^*) \|
\]
which we can rewrite as
\[
\bb{E}_{I^j_\K} \| \Tilde{x}^{j+1} - x^* \| \leq [(\frac{\delta}{1-\lambda})^S + \lambda^j(\frac{1-\lambda}{1-\lambda-\delta})] \|x^0 - x^* \| + \frac{\gamma}{1-\lambda-\delta} \bb{E}_{I^j_\K} \| \nabla_I f_{I^j}(x^*) \| \]\\
\end{proof}

\noindent\textbf{Corollary 2.0.}
Let $\delta < 1$, since $\eta \in (\frac{2\alpha-\sqrt{4\alpha^2-3.75\alpha\beta}}{2\alpha\beta}, \frac{2\alpha+\sqrt{4\alpha^2-3.75\alpha\beta}}{2\alpha\beta})$. Therefore, by definition of $\lambda$, we have:
\[
\lambda = 2(\frac{\delta}{1 + c_{\mathcal{T}}} - \sqrt{1-{\alpha_0}^2}). \]
 
\begin{proof}
By contradiction, assume $\lambda \geq 1$. Therefore, 
\setcounter{equation}{0}
\begin{align*}
& \frac{2\delta}{1 + c_{\mathcal{T}}} - 2\sqrt{1-{\alpha_0}^2} \geq 1 \\
& \frac{2\delta}{1 + c_{\mathcal{T}}} \geq 1 + 2\sqrt{1-{\alpha_0}^2} \geq 1 \\
& \frac{2\delta}{1 + c_{\mathcal{T}}} \geq 1 \\
& 2\delta \geq 1 + c_{\mathcal{T}} \geq 2 \\
& \delta \geq 1
\end{align*}
which is a contradiction, since $\delta < 1$ by definition. Therefore, $\lambda < 1$. Note that $\lambda > 0$ by its definition. 
\end{proof}

\noindent \textbf {Corollary 2.1.} 
  To ensure convergence of our algorithm, the learning rate $\eta$, which is a constant, should be chosen within the range $( \frac{2\alpha - \sqrt{4\alpha^2 - 3.75\alpha\beta}}{2\alpha\beta}, \frac{2\alpha + \sqrt{4\alpha^2 - 3.75\alpha\beta}}{2\alpha\beta})$. For this range to be valid, the following inequality must hold:
\[
\frac{\delta}{1-\lambda} < 1.
\]

\begin{proof}
 Assume by contradiction that $\delta \geq 1 - \lambda$.
Then, by definition of $\lambda$, we get
\setcounter{equation}{0}
\begin{align}
\delta \geq 1 - 2(\frac{\delta}{1+c_{\mathcal{T}}} - \sqrt{1-\alpha_0^2}) \geq 1 - \frac{2\delta}{1+c_{\mathcal{T}}} \\
0 \geq 1 - \frac{2\delta}{1+c_{\mathcal{T}}} - \delta = 1 - \frac{\delta(3 + c_{\mathcal{T}})}{1+c_{\mathcal{T}}} \geq 1 - 2\delta 
\end{align}
where Eq. (2) holds true by the definition of $c_{\mathcal{T}}$.
Now, we have the following:
\begin{align*}
1 - 2\delta \leq 0 \\
1 \leq 2\delta \Rightarrow \delta \geq \frac{1}{2}
\end{align*}
Since $\eta \in (\frac{2\alpha-\sqrt{4\alpha^2-3.75\alpha\beta}}{2\alpha\beta}, \frac{2\alpha+\sqrt{4\alpha^2-3.75\alpha\beta}}{2\alpha\beta})$, then $\delta < \frac{1}{2}$. We then have a contradiction. Therefore,
\[
\frac{\delta}{1-\lambda} < 1,  \ \forall \ \eta \in (\frac{2\alpha-\sqrt{4\alpha^2-3.75\alpha\beta}}{2\alpha\beta}, \frac{2\alpha+\sqrt{4\alpha^2-3.75\alpha\beta}}{2\alpha\beta}).
\]
\end{proof}

\section{More Results}
\subsection{Simulation}



\begin{figure}[H]
    \centering
    \includegraphics[width=0.9\linewidth]{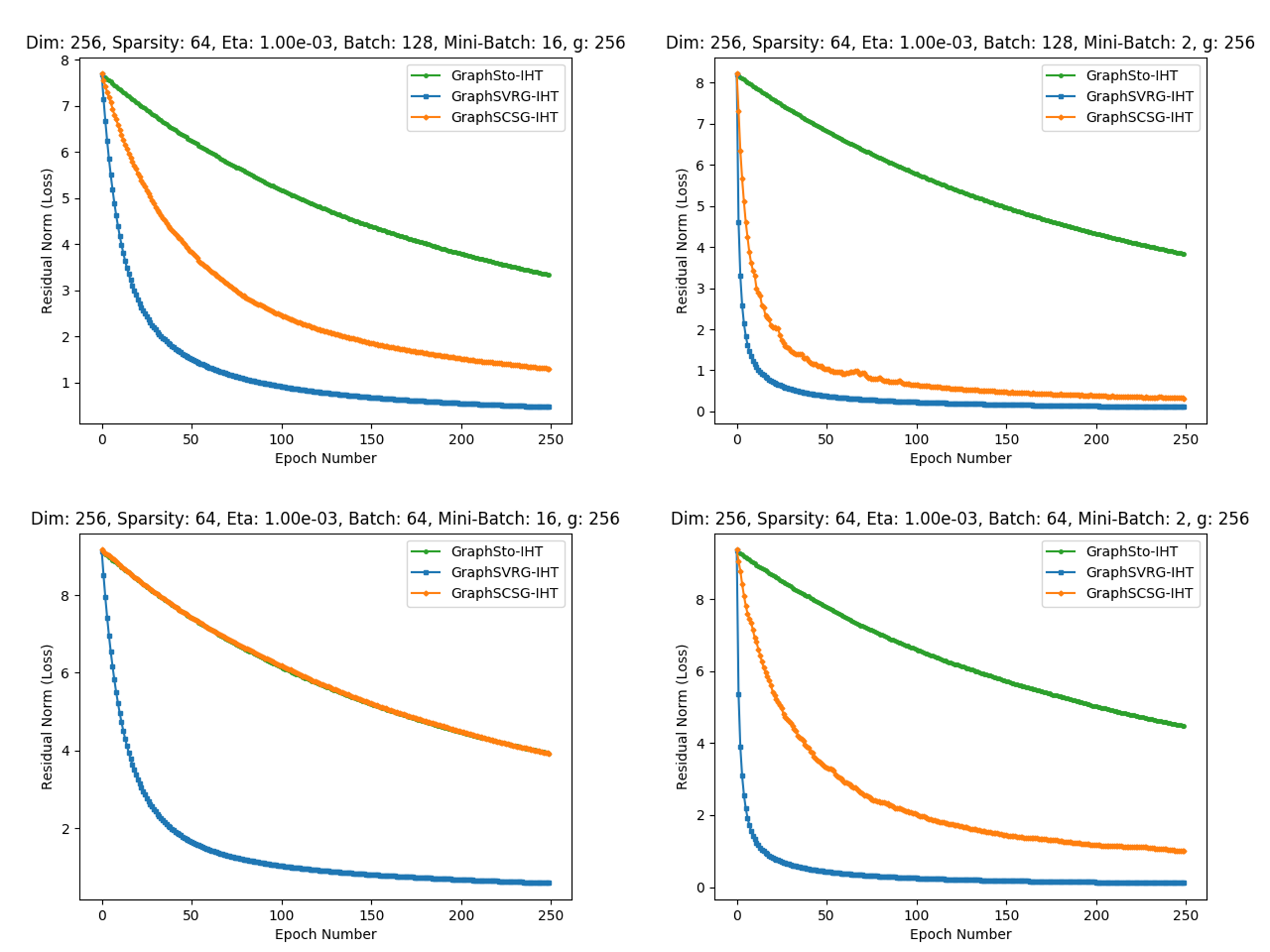}
    \caption{Various choices for $B$ and $b$}
    \label{fig:minibatch}
\end{figure}
\begin{figure}[H]
    \centering
    \begin{tabular}{cc}
    \includegraphics[width=0.45\linewidth]{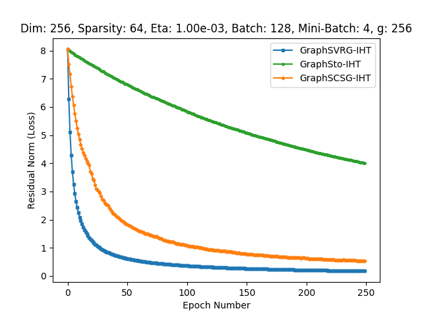} &
    \includegraphics[width=0.45\linewidth]{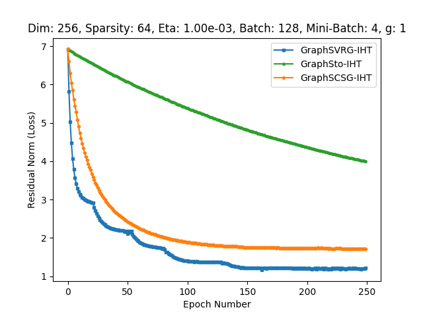}
    \end{tabular}
    \caption{Different number of Connected Components}
    \label{fig:g}
\end{figure}

As pointed out in our main paper, we also explored the effects of mini-batch size in conjunction with batch size. Figure \ref{fig:minibatch} demonstrates the generality of \SCSG. With different settings of $B$ and $b$ \SCSG\ can range in behavior between \STO\ and \SVRG. Not included in the figure is the case when $B$ equals the dimension of the data and $b = 1$, in which case \SCSG\ and \SVRG\ are only distinguished by the number of inner loops (which could be parameterized itself to make the two algorithms identical). 

 We vary the number of connected components, controlled by parameter $g$, in Figure \ref{fig:g}. We observe that when $g$ is low (in this case $g = 1$), the loss function curve does not smoothly decrease but frequently levels off before stepping downward. This is due to the algorithm intermittently finding a more suitable support set, which allows it to more rapidly decrease the objective loss. The restriction of the algorithm to only one connected component makes this phenomenon much more apparent.

\subsection{Breast cancer dataset} All parameters are tuned using 5-fold cross validation, following the experimental settings of \cite{zhou2019stochastic}. The sparsity $s$ is tuned from $[10, 20, ... 100]$ and the block size is tuned from $[n, n/2]$, where $n$ is the number of samples. The task is to find a single connected component, so $g$ is set to 1 for the head and tail projections. The learning rate is tuned using backtracking line search. We record the Balanced Classification Error and Area Under Curve (AUC) scores, as well as the number of nonzeroes for each method over the 20 trials.

\begin{table}[H]
\caption{Balanced Classification Error \tpm \ Std. Dev. on Breast Cancer Dataset}
\label{table:bacc}
\centering
\begin{tabular}{lccccc}
\toprule
           & IHT & \SIHT & \STO & \SVRG & \SCSG \\
\midrule
Trial 1  & 0.352\tpm0.079 & 0.355\tpm0.083 & 0.367\tpm0.091 & \textbf{0.335}\tpm0.048 & 0.474\tpm0.048 \\
Trial 2  & 0.350\tpm0.047 & 0.346\tpm0.045 & 0.395\tpm0.074 &\textbf{0.340}\tpm0.077 & 0.496\tpm0.015 \\
Trial 3  & 0.354\tpm0.060 & 0.370\tpm0.061 & 0.365\tpm0.107 & \textbf{0.318}\tpm0.061 & 0.478\tpm0.061 \\
Trial 4  & 0.351\tpm0.046 & \textbf{0.350}\tpm0.066 & 0.365\tpm0.032 & 0.365\tpm0.032 & 0.459\tpm0.050 \\
Trial 5  & \textbf{0.357}\tpm0.064 & 0.365\tpm0.063 & 0.366\tpm0.065 & 0.368\tpm0.082 & 0.418\tpm0.076 \\
Trial 6  & 0.334\tpm0.050 & \textbf{0.327}\tpm0.046 & 0.330\tpm0.078 & 0.352\tpm0.090 & 0.447\tpm0.071 \\
Trial 7  & 0.343\tpm0.050 & 0.402\tpm0.085 &\textbf{0.321}\tpm0.027 & \textbf{0.321}\tpm0.027 & 0.499\tpm0.004 \\
Trial 8  & 0.352\tpm0.059 & 0.402\tpm0.103 & \textbf{0.338}\tpm0.021 & 0.349\tpm0.030 & 0.489\tpm0.013 \\
Trial 9  & 0.340\tpm0.082 & 0.357\tpm0.058 & \textbf{0.335}\tpm0.110 & 0.343\tpm0.040 & 0.452\tpm0.053 \\
Trial 10 & 0.344\tpm0.068 & 0.351\tpm0.065 & 0.349\tpm0.050 & \textbf{0.331}\tpm0.061 & 0.502\tpm0.005 \\
Trial 11 & 0.339\tpm0.056 & \textbf{0.318}\tpm0.061 & 0.353\tpm0.076 & 0.357\tpm0.054 & 0.473\tpm0.054 \\
Trial 12 & 0.355\tpm0.061 & 0.394\tpm0.053 & 0.349\tpm0.088 & \textbf{0.329}\tpm0.082 & 0.463\tpm0.072 \\
Trial 13 & 0.327\tpm0.053 & 0.317\tpm0.040 & \textbf{0.312}\tpm0.065 & 0.323\tpm0.060 & 0.457\tpm0.051 \\
Trial 14 & 0.362\tpm0.058 & 0.340\tpm0.071 & 0.348\tpm0.019 &\textbf {0.330}\tpm0.054 & 0.453\tpm0.045 \\
Trial 15 & 0.381\tpm0.048 & 0.346\tpm0.051 & \textbf{0.297}\tpm0.059 & \textbf{0.297}\tpm0.059 & 0.416\tpm0.069 \\
Trial 16 & 0.330\tpm0.094 & \textbf{0.320}\tpm0.089 & 0.330\tpm0.074 & 0.339\tpm0.064 & 0.488\tpm0.055 \\
Trial 17 & 0.349\tpm0.074 & 0.332\tpm0.067 & 0.353\tpm0.078 & \textbf{0.330}\tpm0.056 & 0.437\tpm0.055 \\
Trial 18 & 0.331\tpm0.096 & 0.331\tpm0.096 & 0.292\tpm0.067 & \textbf{0.290}\tpm0.069 & 0.462\tpm0.053 \\
Trial 19 & 0.327\tpm0.052 & 0.333\tpm0.054 & 0.311\tpm0.021 & \textbf{0.306}\tpm0.023 & 0.476\tpm0.017 \\
Trial 20 & \textbf{0.333}\tpm0.038 & 0.338\tpm0.026 & 0.352\tpm0.064 & 0.376\tpm0.036 & 0.427\tpm0.052 \\
\midrule
Averaged & 0.345\tpm0.065 & 0.350\tpm0.072 & 0.341\tpm0.073 & \textbf {0.335}\tpm0.062 & {0.463\tpm0.057} \\
\bottomrule
\end{tabular}
\end{table}

\begin{table}
\caption{AUC score \tpm \ Std. Dev. on Breast Cancer Dataset}
\label{table:auc}
\centering
\begin{tabular}{lccccc}
\toprule
            & IHT & \SIHT & \STO & \SVRG & \SCSG \\
\midrule
Trial 1  & 0.726\tpm0.067 & \textbf{0.736}\tpm0.054 & 0.720\tpm0.030 & 0.729\tpm0.039 & 0.713\tpm0.023 \\
Trial 2  & 0.683\tpm0.067 & 0.692\tpm0.064 & 0.697\tpm0.044 & \textbf{0.712}\tpm0.062 & 0.696\tpm0.101 \\
Trial 3  & 0.716\tpm0.062 & \textbf{0.726}\tpm0.077 & \textbf{0.726}\tpm0.064 & 0.715\tpm0.065 & 0.703\tpm0.027 \\
Trial 4  & 0.695\tpm0.059 & \textbf{0.701}\tpm0.061 & 0.687\tpm0.041 & 0.687\tpm0.041 & 0.693\tpm0.033 \\
Trial 5  & 0.699\tpm0.042 & 0.700\tpm0.067 & \textbf{0.707}\tpm0.040 & 0.701\tpm0.028 & 0.701\tpm0.045 \\
Trial 6  & 0.677\tpm0.071 & 0.673\tpm0.074 & \textbf{0.704}\tpm0.051 & 0.669\tpm0.071 & 0.640\tpm0.101 \\
Trial 7  & 0.712\tpm0.069 & 0.719\tpm0.066 & \textbf{0.741}\tpm0.055 & \textbf{0.741}\tpm0.055 & 0.701\tpm0.069 \\
Trial 8  & 0.711\tpm0.081 & 0.707\tpm0.055 & \textbf{0.721}\tpm0.062 & 0.717\tpm0.055 & 0.657\tpm0.075 \\
Trial 9  & 0.717\tpm0.066 & 0.718\tpm0.068 & \textbf{0.723}\tpm0.044 & 0.718\tpm0.026 & 0.719\tpm0.058 \\
Trial 10 & 0.710\tpm0.060 & \textbf{0.711}\tpm0.060 & 0.691\tpm0.052 & 0.702\tpm0.062 & 0.655\tpm0.103 \\
Trial 11 & 0.708\tpm0.082 & 0.719\tpm0.086 & \textbf{0.722}\tpm0.085 & 0.711\tpm0.074 & 0.693\tpm0.142 \\
Trial 12 & 0.713\tpm0.025 & 0.711\tpm0.045 & \textbf{0.736}\tpm0.055 & 0.730\tpm0.060 & 0.664\tpm0.057 \\
Trial 13 & 0.714\tpm0.058 & \textbf{0.719}\tpm0.058 & 0.718\tpm0.073 & 0.703\tpm0.071 & 0.704\tpm0.074 \\
Trial 14 & 0.700\tpm0.058 & 0.705\tpm0.067 & 0.711\tpm0.064 & \textbf{0.712}\tpm0.046 & 0.665\tpm0.061 \\
Trial 15 & 0.696\tpm0.033 & 0.687\tpm0.040 & \textbf{0.721}\tpm0.058 & \textbf{0.721}\tpm0.058 & 0.715\tpm0.040 \\
Trial 16 & \textbf{0.729}\tpm0.081 & 0.725\tpm0.083 & 0.723\tpm0.072 & 0.725\tpm0.073 & 0.734\tpm0.049 \\
Trial 17 & 0.720\tpm0.055 & \textbf{0.722}\tpm0.062 & 0.718\tpm0.037 & 0.717\tpm0.037 & 0.681\tpm0.061 \\
Trial 18 & 0.712\tpm0.039 & 0.712\tpm0.039 & \textbf{0.733}\tpm0.036 & 0.730\tpm0.034 & 0.679\tpm0.056 \\
Trial 19 & 0.721\tpm0.067 & 0.724\tpm0.068 & 0.736\tpm0.056 & \textbf{0.738}\tpm0.054 & 0.701\tpm0.057 \\
Trial 20 & \textbf{0.720}\tpm0.031 & 0.706\tpm0.027 & 0.717\tpm0.045 & 0.705\tpm0.033 & 0.703\tpm0.056 \\
\midrule
Averaged & 0.709\tpm0.062 & 0.711\tpm0.064 & \textbf{0.717}\tpm0.057 & 0.714\tpm0.057 & 0.691\tpm0.074 \\
\bottomrule
\end{tabular}
\end{table}

\begin{table}
\caption{Num. nonzeroes \tpm \ Std. Dev. on Breast Cancer Dataset}
\label{table:nonzero}
\centering
\begin{tabular}{lccccc}
\toprule
            & IHT & \SIHT & \STO & \SVRG & \SCSG \\
\midrule
Trial 1  & 082.0\tpm36.00 & 066.0\tpm28.00 & \textbf{033.4}\tpm10.38 & 033.8\tpm17.31 & 064.0\tpm17.72 \\
Trial 2  & 044.0\tpm08.00 & \textbf{038.0}\tpm13.27 & 061.2\tpm41.81 & 079.0\tpm32.00 & 087.2\tpm04.12 \\
Trial 3  & 094.0\tpm12.00 & 068.0\tpm31.87 & \textbf{030.4}\tpm12.97 & 040.2\tpm17.03 & 086.2\tpm13.33 \\
Trial 4  & 074.0\tpm12.00 & 062.0\tpm31.87 & \textbf{051.6}\tpm08.33 & \textbf{051.6}\tpm08.33 & 075.0\tpm18.17 \\
Trial 5  & 048.0\tpm04.00 & \textbf{036.0}\tpm13.56 & 054.2\tpm14.37 & 057.6\tpm15.99 & 078.2\tpm12.69 \\
Trial 6  & 090.0\tpm00.00 & 080.0\tpm20.00 & \textbf{041.0}\tpm12.44 & 047.4\tpm16.21 & 094.6\tpm09.46 \\
Trial 7  & \textbf{010.0}\tpm00.00 & 048.0\tpm39.19 & 079.8\tpm09.60 & 079.8\tpm09.60 & 091.4\tpm06.77 \\
Trial 8  & \textbf{040.0}\tpm00.00 & 048.0\tpm29.26 & 083.0\tpm29.26 & 072.0\tpm36.55 & 063.0\tpm19.13 \\
Trial 9  & 092.0\tpm04.00 & 076.0\tpm33.23 & 035.2\tpm13.50 & \textbf{019.8}\tpm07.49 & 078.0\tpm18.06 \\
Trial 10 & \textbf{050.0}\tpm00.00 & 052.0\tpm04.00 & 053.0\tpm13.64 & 052.0\tpm14.00 & 088.2\tpm11.53 \\
Trial 11 & 074.0\tpm08.00 & \textbf{054.0}\tpm20.59 & 059.0\tpm23.39 & 073.0\tpm22.49 & 078.6\tpm20.34 \\
Trial 12 & 056.0\tpm08.00 & \textbf{022.0}\tpm09.80 & 095.2\tpm10.80 & 080.8\tpm31.52 & 083.4\tpm15.42 \\
Trial 13 & \textbf{046.0}\tpm12.00 & 052.0\tpm21.35 & 067.0\tpm22.27 & 067.0\tpm23.58 & 087.4\tpm09.73 \\
Trial 14 & 078.0\tpm24.00 & 058.0\tpm23.15 & \textbf{028.8}\tpm06.76 & 030.4\tpm06.65 & 074.2\tpm16.27 \\
Trial 15 & 082.0\tpm04.00 & 052.2\tpm27.63 & \textbf{045.0}\tpm00.00 & \textbf{045.0}\tpm00.00 & 088.0\tpm14.35 \\
Trial 16 & 052.0\tpm16.00 & 056.0\tpm19.60 & 047.0\tpm16.91 & \textbf{041.2}\tpm24.55 & 073.6\tpm23.69 \\
Trial 17 & 092.0\tpm04.00 & 064.0\tpm23.32 & \textbf{024.2}\tpm07.52 & 027.0\tpm06.51 & 084.4\tpm11.55 \\
Trial 18 & 094.0\tpm12.00 & 094.0\tpm12.00 & 044.2\tpm17.90 & \textbf{034.2}\tpm02.14 & 082.0\tpm14.35 \\
Trial 19 & 100.0\tpm00.00 & \textbf{030.0}\tpm20.98 & 043.6\tpm08.28 & 038.6\tpm03.83 & 087.2\tpm05.42 \\
Trial 20 & 020.0\tpm00.00 & 050.0\tpm26.83 & \textbf{042.2}\tpm20.43 & 049.8\tpm25.20 & 088.2\tpm12.92 \\
\midrule
Averaged & 065.9\tpm28.36 & 055.3\tpm29.26 & \textbf{051.0}\tpm25.39 & \textbf{051.0}\tpm26.44 & 081.6\tpm16.80 \\
\bottomrule
\end{tabular}
\end{table}

\end{document}